%% file: timing.tex
\documentclass{sig-alternate-05-2015}

\usepackage[sc]{mathpazo}
\usepackage[T1]{fontenc}

\PassOptionsToPackage{hyphens}{url}

\usepackage{hyperref}
\usepackage[usenames,dvipsnames]{color}
\definecolor{dark-gray}{gray}{0.30}
\definecolor{orange}{rgb}{0.8,0.4,0}
\definecolor{mylink}{RGB}{18,68,115}

\hypersetup{letterpaper,bookmarksopen,bookmarksnumbered,
pdfpagemode=UseOutlines,
colorlinks=true,
linkcolor=mylink,
anchorcolor=blue,
citecolor=mylink,
filecolor=blue,
menucolor=blue,
urlcolor=mylink,
}

\usepackage[usenames,dvipsnames]{color}
\usepackage{amssymb}
\usepackage{booktabs}
\usepackage{graphicx}
\usepackage{subfigure}
\usepackage[normalem]{ulem}
\usepackage[keeplastbox]{flushend}

\newcommand{\norm}[1]{\left\lVert#1\right\rVert}

\input{vars}

\usepackage[labelfont=bf,font=small]{caption}
\DeclareCaptionType{copyrightbox}

\begin{document}
\setcopyright{acmlicensed}
\conferenceinfo{HRI '17,}{March 6--9, 2017, Vienna, Austria.}
\isbn{978-1-4503-4336-7/17/03}\acmPrice{\$15.00}
\doi{http://dx.doi.org/10.1145/2909824.3020221}
\clubpenalty=10000
\widowpenalty = 10000

\newcommand{\eref}[1]{(\ref{#1})}
\newcommand{\sref}[1]{Sec. \ref{#1}}
\newcommand{\figref}[1]{Fig.\ref{#1}}

\newcommand{\adnote}[1]%
 {\textcolor{blue}{\textbf{AD: #1}}}

\newcommand{\dhmnote}[1]%
 {\textcolor{green}{\textbf{DHM: #1}}}

\newcommand{\aznote}[1]%
 {\textcolor{red}{\textbf{AZ: #1}}}

\newcommand{\prg}[1]{\noindent\textbf{#1. }} 
\newcommand{\be}[1]{\textbf{\emph{#1}}} 
 
\title{Expressive Robot Motion Timing}

\numberofauthors{4}
\author{
  \alignauthor Allan Zhou \\
  \affaddr{UC Berkeley}
  \alignauthor Dylan Hadfield-Menell \\
  \affaddr{UC Berkeley}
  \alignauthor Anusha Nagabandi \\
  \affaddr{UC Berkeley}
  \and
  \alignauthor Anca D. Dragan \\
  \affaddr{UC Berkeley}
}

\maketitle
\begin{abstract}
  Our goal is to enable robots to \emph{time} their motion in a way that is
  purposefully expressive of their internal states, making them more
  transparent to people. We start by investigating what types of
  states motion timing is capable of expressing, focusing on robot
  manipulation and keeping the path constant while systematically
  varying the timing. We find that users naturally pick up on certain
  properties of the robot (like confidence), of the motion (like
  naturalness), or of the task (like the weight of the object that the
  robot is carrying). We then conduct a hypothesis-driven experiment
  to tease out the directions and magnitudes of these effects, and use
  our findings to develop candidate mathematical models for how users
  make these inferences from the timing. We find a strong correlation
  between the models and real user data, suggesting that robots can
  leverage these models to autonomously optimize the timing of their
  motion to be expressive.
\end{abstract}


\input{introduction}
\input{exploratory}
\input{experiment}
\input{models}

\input{discussion}

\bibliographystyle{plain}
\bibliography{references}

\end{document}

%% file: vars.tex
\newcommand{\traj}{\ensuremath{\mathbf{q}}}
\newcommand{\ind}{\ensuremath{i}}
\newcommand{\traji}{\ensuremath{q_\ind}}
\newcommand{\dur}{\ensuremath{T}}
\newcommand{\duration}{\ensuremath{T_{N}}}
\newcommand{\ts}{\ensuremath{T_\ind}}
\newcommand{\T}{\ensuremath{\mathbf{T}}}
\newcommand{\vel}{\ensuremath{v_\ind}}
\newcommand{\velEE}{\ensuremath{v_\ind^{EE}}}

\newenvironment{tightlist}%
{\begin{list}{$\bullet$}{%
    \setlength{\topsep}{0in}
    \setlength{\partopsep}{0in}
    \setlength{\itemsep}{0in}
    \setlength{\parsep}{0in}
    \setlength{\leftmargin}{1.5em}
    \setlength{\rightmargin}{0in}
    \setlength{\itemindent}{.2in}
}
}%
{\end{list}
}

%% file: introduction.tex
\section{Introduction}
Robot motion trajectories have two components. There is a kinematic
component, which is the geometric path through the robot's
configurations space -- a sequence of configurations that the robot
will traverse. But there is also a timing component -- a function that
assigns a time stamp to each configuration along the path, dictating
\emph{how} the robot will traverse the configuration sequence.

\begin{figure}[t]
  \includegraphics[width=\columnwidth]{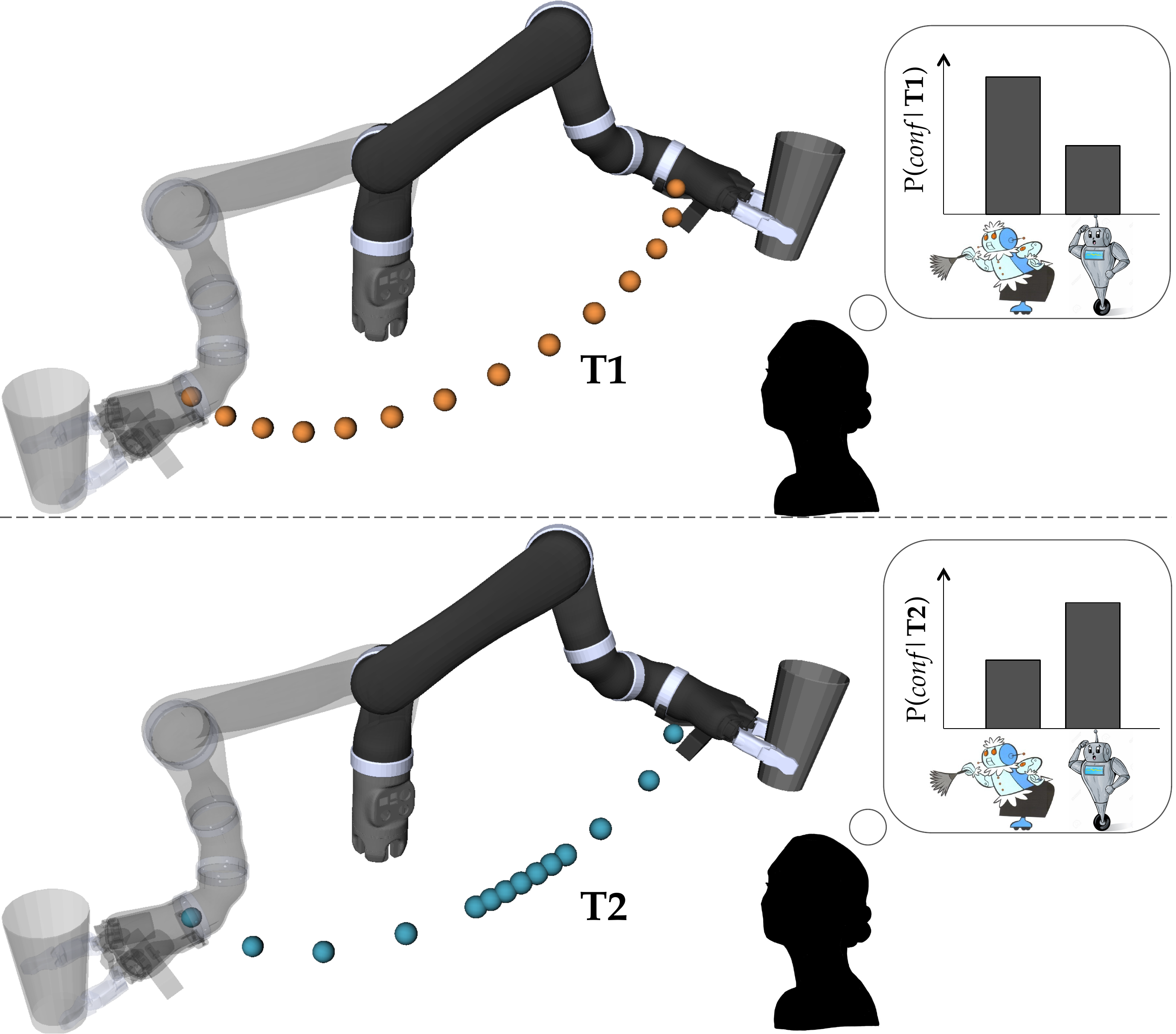}

  \caption{Different timings of the same motion convey different things about the robot. 
  We find effects on perceived confidence, naturalness, even the perceived weight of 
  the object being manipulated. We develop mathematical models for these perceptions
  that correlate with user data and enable robots to optimize their timing for expressiveness.}
  \label{fig:front}
\end{figure}

Robotics motion planners for manipulation tend to focus on the path
\cite{lavalle1998rapidly, lavalle2001randomized, zucker2013chomp,
  schulman2013finding}, with few exceptions \emph{explicitly}
incorporating timing, for instance to improve efficiency or conservative obstacle
avoidance \cite{byravan2014space, bobrow1985time}. Most commonly,
timing is an after-thought in robotics, left to the controller to
assign post-hoc.

And yet, timing is crucial in \emph{HRI}. Imagine seeing a robot arm
carry a cup smoothly across the table, like in the top image in \figref{fig:front}. 
Now, imagine seeing a different arm pausing
and restarting, slowing down and then speeding back up, like in the bottom image. 
The path
might be the same, but the difference in timing might make us 
think very differently about the robots and about what they are doing.
We might think that the second robot is less capable, or maybe that
its task is more difficult. Perhaps it doesn't have as much payload, 
perhaps the cup is heavier, or perhaps it does not know what to do: 
\begin{quote}
  \emph{The timing of a path affects how observers perceive the robot
    and the task that it is performing.}
\end{quote}

Studies have already shown that
the average velocity and changes in velocity of motion affect
perceptions of expressed emotion \cite{camurri2003recognizing},
intent \cite{gielniak2011generating},
elation \cite{bethel2008survey},
animacy \cite{tremoulet2000perception}, arousal and dominance
\cite{saerbeck2010perception}, and energy
\cite{blythe1999motion}. When it comes to robot
motion, human observers will interpret the timing regardless of
whether the robot is planning to express anything or not. Our goal is
to give robots control over what their timing inadvertently expresses:

\begin{quote}
  \emph{Robots should leverage timing
    to be more expressive of their internal states.}
\end{quote}

Techniques from animation can be useful in improving robot expressiveness
\cite{takayama2011expressing,ribeiro2012illusion}, and animated characters
have long taken advantage of
timing, both for making motion more natural (e.g. ease-in ease-out is
one of the 12 animation principles \cite{thomas1981disney}), and
more expressive \cite{owen1999timing, saerbeck2010perception}.
This made timing a center of focus in the graphics community,
developing automated tools for assigning timing to a
path. Most tools still leave the \emph{animator} in control of the
timing, but simplify the assignment process (e.g. by allowing
the animator to ``act out'' the timing of a motion with something
like a pen and tablet \cite{Terra:2004:PTK:1028523.1028556}). Other
tools \emph{align} timing to a different trajectory or an external
event like a beat \cite{Kim:2003:RSB:1201775.882283, hsu2005style}. Others yet
\emph{re-time} a particular motion to satisfy new constraints (like
finishing faster) while maintaining physical \emph{realism}
\cite{mccann2006physics}. 

Overall, although realistic timing can be automated, 
even virtual characters still rely
on an external expert when it comes to \emph{expressive} timing --
be it on an animator or on an artist's trajectory. Robots, on the other
hand, can't afford to rely on experts for every motion they need to perform. 
They plan their motion autonomously, and have to autonomously 
decide on how to time it. 

Our focus is on enabling robots to produce expressive timing.
Two questions remain in this area. First, there
is the question of what timing \emph{can} express in the first
place -- prior work looked at effects on perceived emotional state, but 
are there also effects on function-related properties? 
Second, there is the synthesis question -- how can we enable
robots to autonomously \emph{generate} timing \emph{from scratch} that
is purposefully expressive, rather than efficient or physically
realistic. We take a step in this direction by analyzing motion timing
during manipulation, from an open ended study, to hypothesis-driven
experiments, to candidate mathematical models that capture human
timing-based inferences.

We make three contributions:

\prg{Exploring the possible effects of timing} So far, studies
focusing on timing mainly looked for effects on perceived emotional state. 
We designed and conducted a study to identify what types
of variables timing influences more broadly. Rather than biasing users
with questionnaires that already suggest how the timing
should be interpreted, we used simple open-ended questions. We systematically
manipulated timing across three axes inspired by prior work in a
factorial design, and asked users to characterize the robot and the
task. We clustered their responses to identify common interpretations,
and uncovered robot competence,
confidence, disposition, along with (unsurprisingly) motion naturalness, and a 
manipulation-specific characteristic of the task: the
weight of the object being manipulated by the robot. This list of
variables by no means comprise the entirety of timing effects, nor is
it as specific as we would ultimately desire. It does, however,
provide us with a rich set of dependent measures for more in-depth
analysis.

\prg{Experiments that test these effects} Only after identifying
candidate dependent variables based on open-ended questions
did we put these effects to the test. We conducted a hypothesis-driven experiment to
understand the magnitude and directionality for each. 
Some of our findings
support intuition, like the robot being perceived as less confident if
it pauses during the motion. Others are quite surprising. For
instance, when the robot is carrying an object, we found that people
estimated that object to have approximately the same weight regardless
of whether or not the timing had pauses or speed changes. Overall, pausing had 
a much stronger effect than speed. 

\prg{Mathematical models and evaluation} 
Our experiment shows \emph{what} effects timing has, but not \emph{why}.
For robots to generate their timing autonomously in different situations,
they need a mechanism for generalizing these findings. 
We attempt such a mechanism for three of the dependent variables.
We introduce mathematical models for the inferences that humans
make from motion timing. We take a Bayesian inference approach,
in which the timing serves as an observation to the human
about the states that they can't observe, like the robot's confidence
or the object's weight. We show strong correlations between these models
and the real user data. The models are constructive, in the sense
that robots can use them to optimize their timing to be expressive.

Overall, this paper shows how several timing features interact to
affect perceptions of the robot and task, and uses these findings to
introduce optimization criteria that correlate with the user data and
that robots could use to autonomously \emph{time} motion in a way that
is expressive (e.g. of the robot's confidence). We look forward to
future work on further analysis and refinement of these criteria to ensure generality
across settings, as well as further exploration of effects that are
more difficult to model, such as how timing influences the robot's
perceived disposition.

%% file: exploratory.tex
\begin{figure*}
\centering
\includegraphics[width=\textwidth]{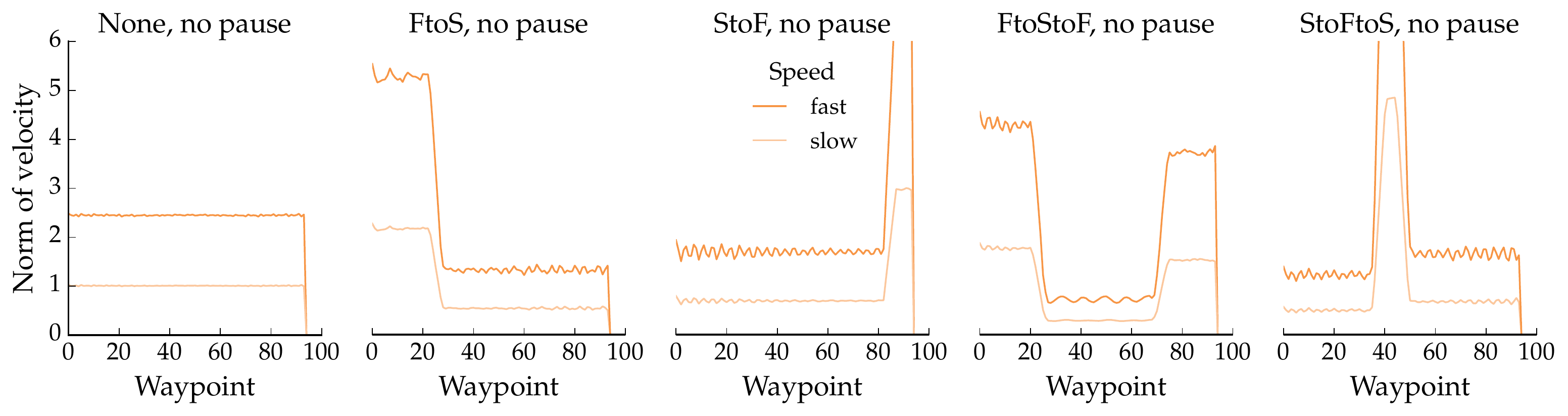}\\
\includegraphics[width=\textwidth]{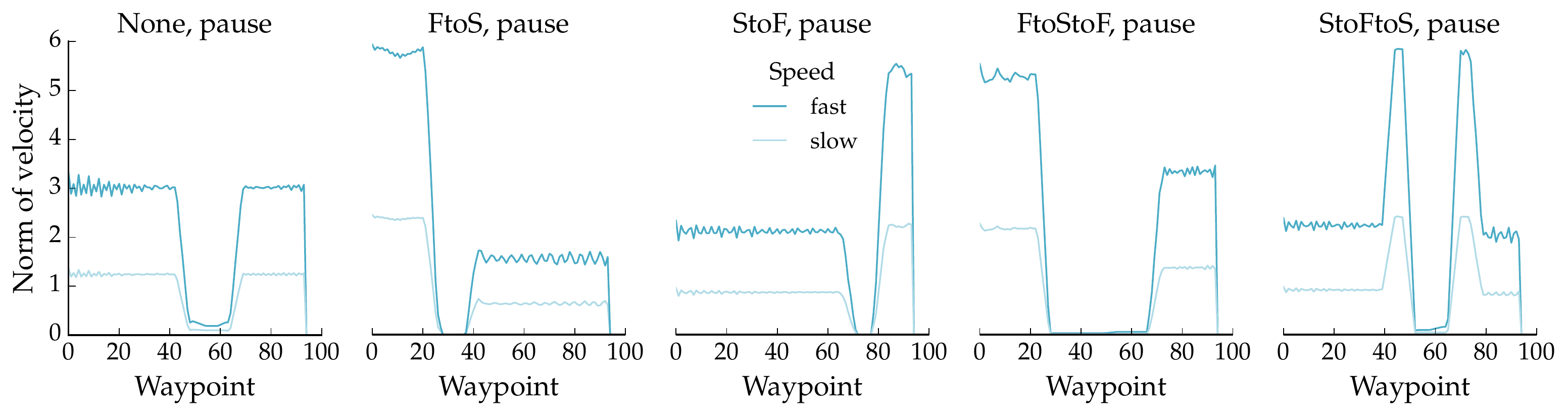}
\caption{Norm of the robot's configuration space velocity $\vel$ for each way point configuration $\traji$ in each of our 20 conditions. Each column is a different speed change pattern, with the top row representing the conditions without a pause and the bottom representing the conditions with a pause, where we see the velocity go to 0. Each plot contains both the slow motion (lighter color) and its fast counterpart (darker color).}
\label{fig:velocity}
\end{figure*}

\section{Notation}

A trajectory in our experiment consists of two components: the
sequence of  of way point configurations that the robot moves through, and the
time at which the robot reaches each way point. We use \traji{} to represent the
$\ind^{\text{th}}$ robot configuration. \ts{} is the time that the
robot reaches the $\ind^{\text{th}}$ configuration. We assume that
trajectories begin at time 0. \duration{} is the total duration of the
trajectory (i.e., the time at which the robot reaches the final
configuration). 

We will usually be interested in the speed the robot
travels over the course of its trajectory. We use $$\vel =
\frac{q_{i+1} - \traji}{\dur_{i+1} - \ts}$$ to represent this
velocity (in radians per second). 
We use $$\velEE =
\frac{\phi(q_{i+1}) - \phi(\traji)}{\dur_{i+1} - \ts}$$
where $\phi$ is the robot's forward kinematics function, to represent the 
velocity of the end effector (in meters per second).  

To summarize:
\begin{tightlist}
\item[\traj:] A sequence of robot configurations that represents the kinematic component (path) of a trajectory.
\item[\traji:] The $\ind^{\text{th}}$ robot configuration in the trajectory.
\item[\T:] A sequence of time stamps that represents the timing component of a trajectory.
\item[\ts:] The time when the robot reaches \traji.
\item[\duration:] The total time taken by the trajectory.
\item[\vel:] The velocity of the robot from \traji{} to $q_{i+1}$.
\item[\velEE:] The end effector velocity of the robot from \traji{} to $q_{i+1}$.
\end{tightlist}

\section{Exploratory Study}
\label{sec:exploratory}
We start with a study that builds on prior work to find what kinds of
effects timing can have on what people infer about the robot during a
manipulation task. Our goal with this study is to find the different
dimensions of perception that timing affects, i.e. the dependent
measures we should test -- is it energy, elation, dominance, or
something different? We need to avoid biasing the users towards a
particular interpretation, so we ask the users open-ended questions
and use their responses to form hypotheses for our next experiment.

\subsection{Study Design}
\label{sec:exploratory-design}
\prg{Robot Task} We used a Kinova 6DOF Mico arm (\figref{fig:front}) in
our study.  We chose one of the most common interactive manipulation
tasks for the robot: a handover
\cite{cakmak2011using, strabala2013towards, mainprice2011planning, moon2014meet}.
The robot carried an object (a cup) from a table to a handover configuration
(see \figref{fig:front}).

\begin{figure*}
\includegraphics[width=\textwidth]{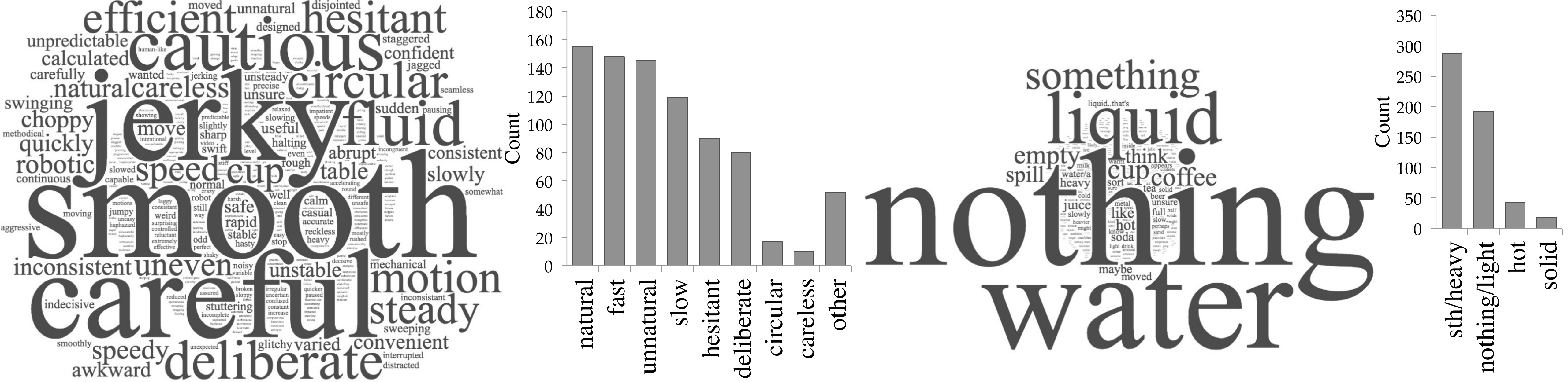}
\caption{Words that participants in the exploratory study used to
  characterize the robot (left) and what it was carrying (right).
  The histograms cluster words into equivalence classes.
  We use these classes to devise the 5 dependent
  measures for our experiment, 4 of which are perceptions
  of the robot and its motion: competence, confidence, disposition,
  and naturalness. The other dependent measure is functional, the
  perceived weight of the object in the handover.
}
\label{fig:exploratory}
\end{figure*}
\prg{Manipulated Factors} The biggest challenge in studying the
effects of timing on people's perceptions is identifying which
timing variations to test. A simple answer would be to randomly sample
timings, which would uniformly cover the space of all
timings. However, they would almost uniformly be interpreted in the
same way -- as erratic and unnatural. Instead, we decided to
\emph{systematically} generate timings by manipulating several factors.

Our first factor is overall \emph{\textbf{speed}}. Previous studies
found that overall robot speed has effects on perceptions
\cite{bethel2008survey, saerbeck2010perception}.
We use 2 levels for this factor: \emph{slow} and
\emph{fast}.

Our second factor is \emph{\textbf{change in speed}}. In studies on
abstract characters and human motion, changes in speed have been shown
to affect perceived animacy \cite{tremoulet2000perception},
emotion content \cite{camurri2003recognizing}, and energy
\cite{blythe1999motion}. Here, we considered 0, 1, and 2 changes,
leading to a total of 5 levels for this factor: \emph{none} (no
change), \emph{StoF} (starting to go faster), \emph{FtoS} (slowing
down), \emph{StoFtoS} (faster, then slower), and \emph{FtoStoF}
(slower, than faster). \figref{fig:velocity} (top) shows the magnitude of the velocity $\vel$ across the
trajectory way points for each of these patterns, and for both the overall slow and the overall fast levels.

Finally, we also explore an edge case of change in speed: coming to a
full stop.  Our third factor is
thus \emph{\textbf{pause}}, with 2 levels: either the robot pauses or
it does not. The pausing variants are at the bottom of \figref{fig:velocity}, and 
they differ in that the velocity goes to 0 for a portion of the trajectory. 
 
We used a 2 by 5 by 2 factorial design, leading to a total of 20 conditions, each corresponding to a different timing $\T$ for a path $\traj$ (shown in \figref{fig:velocity}).

\prg{Dependent Measures} We asked users to describe how the robot
moved the cup, what adjectives they would use to characterize the
robot, and what they think is in the cup.

\prg{Subject Allocation} We wanted a within-subjects design to enable
users to see multiple possible timings and have bases for comparisons,
as they would if they would interact with the robot on a longer
term. However, we had 20 conditions, making within-subjects
infeasible. We opted for a randomized assignment, where each
participant evaluated 8 randomly sampled conditions. There were a
total of 61 participants (63\% male and 37\% female, median age $=36$)
all from the United States and recruited
through Amazon's Mechanical Turk platform. All had a
minimum approval rating of 95\% on Mechanical Turk.

\subsection{Analysis}

We started by computing word counts for each
question. \figref{fig:exploratory} shows the word cloud that this
induced for the question of describing the robot, and identifying what the
robot is carrying.\footnote{\scriptsize We quickly realized that when asked to describe how
the robot moved the cup, users quite literally described what the
robot did (e.g. ``move the cup slowly away from the table''), and we
are not including the analysis for that question.} We then clustered
the words into equivalence classes for easier analysis, ignoring words
that appear fewer than 3 times in the data.

\prg{Adjectives}
Two of the most common adjectives used to describe the robot
were literal: \emph{slow} and \emph{fast} (or quick,
speedy, rapid, efficient), giving rise to two of our clusters. But
beyond those, many users described the motion as smooth, natural,
predictable, or fluid, which formed the \emph{natural} cluster with
the highest word count (left histogram in
\figref{fig:exploratory}). The counterparts were also present
(unnatural, jerky, robotic, mechanical, uneven, awkward), forming the
\emph{unnatural} cluster.  And finally, users described the robot as
careful, cautious, deliberate, hesitant, indecisive, calculated. We
split these into two clusters: one that suggests low confidence but
high ability, like \emph{deliberate}, and one that suggests low
confidence and low ability, like \emph{hesitant}. The exact split of
these is difficult to determine, so the relative counts for deliberate and hesitant
should be taken with a grain of salt. The sum of the two, however, is 
important, and is larger than any of our other clusters, suggesting 
the importance of timing in perceptions of competence/confidence.
Their
counterpart, \emph{careless}, is also present. The histogram plots
clusters with more than 10 entries, and groups the remaining words
into ``other''.

Based on these clusters, we see that motion timing might
affect perceived motion \be{naturalness}, but also two important other
variables: perceived \be{competence} and perceived
\be{confidence}. Confidence alone is not sufficient, because it
doesn't enable us to differentiate between hesitation and
deliberation. But taken together, these two variables can help
represent our clusters. Surprisingly, none of the descriptions
directly related to arousal, dominance, energy (with the exception of
the word ``aggressive''). Previous studies found effects when directly
measuring these, but they do not seem to specifically occur when users
aren't directly asked about them. However, the adjectives that
participants used could be interpreted as suggesting higher or lower
values for these variables, and we capture them with a  broader term: 
perceived \be{disposition} (with positive or negative values).

\prg{Object} For what the robot was carrying, the most common words were ``water''
and ``nothing''. While this is not surprising, their difference is
important: the cup is heavier when it has something
inside. Participants provided many other options, and we
differentiated them between standard (e.g. something, water, soda,
etc.), options that mentioned hot liquids (e.g. coffee), and solids
that cannot be spilled, e.g. (jell-o, or even metal). Hot
contents and solid contents were uncommon (right histogram) and not
generalizable far beyond open containers, so we identify one variable
here: perceived \be{weight} of the object that the robot is carrying.

%% file: experiment.tex
\section{Experiments on Timing Effects}
\label{sec-experiment}

We designed our experiments based on the findings from the exploratory
study. The word counts already suggest certain effects, e.g. that
slower motion tends to be more often described as careful or cautious
or deliberate than fast motion (all three appear in top 15 words for
slow and do not for fast). 

However, we noticed that the difference
remains the same when considering changes in speed and when
not. Because of this and because changes in speed and pauses are
related, we decided to separate into two experiments rather than
testing all possible interactions: a first one focusing on speed and
pauses, and a second focusing on the speed change patterns.
\begin{figure*}[t]
  \includegraphics[width=\textwidth]{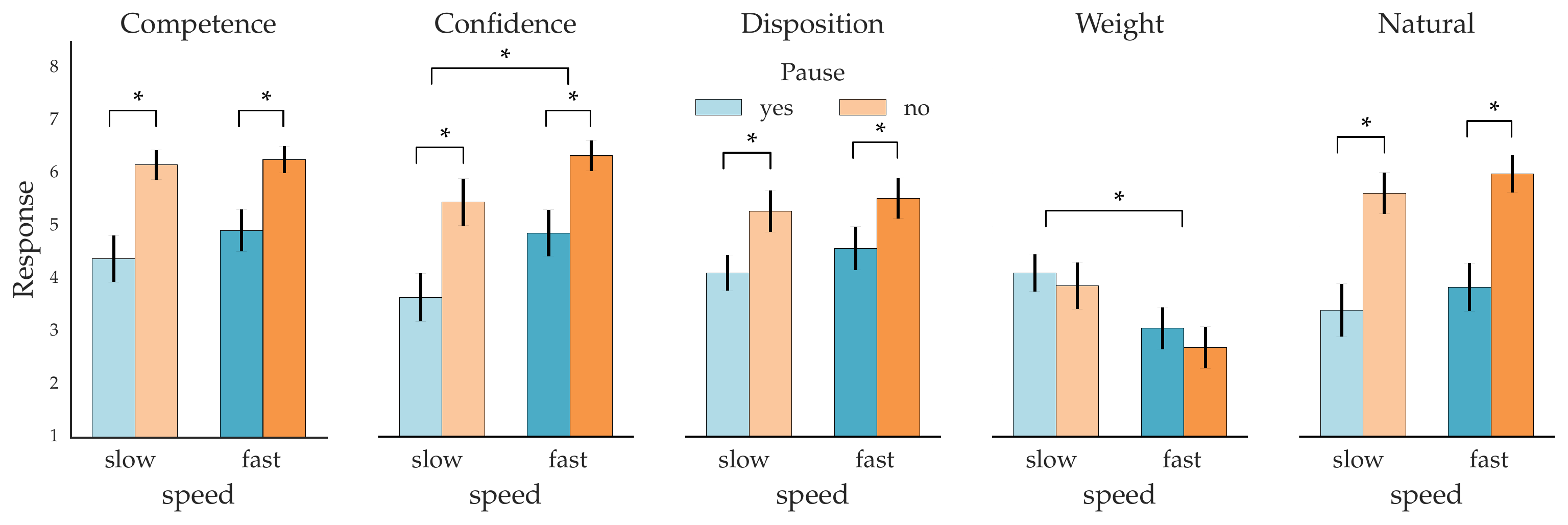}\\
  \caption{ Our first hypothesis-driven experiment measured the effect
    of overall speed and pausing and their interaction. Speed
    significantly affected perceived confidence and weight.  Pausing
    significantly affected perception of every property \emph{except}
    weight. }
  \label{fig:results_exp2}
\end{figure*}

\subsection{Speed, Pauses, and Their Interaction}
\label{sec:speedpause}
\subsubsection{Experiment Design}
\prg{Manipulated Factors} In this experiment, we manipulate the
\be{speed} and \be{pause} factors as we did in \sref{sec:exploratory-design}.

\prg{Dependent Measures} We use the measures we identified 
in the previous section. We measure perceived \be{competence,
  confidence, disposition, naturalness}, and \be{weight} using 7-point
 scales. 

For each dependent measure, the scales were
labeled at either end and in the very middle. For example, the
disposition scale was labeled ``very negative'',
``neither positive nor negative'', and ``very positive'' on the leftmost,
middle, and rightmost options, respectively. We chose to label the
 scales in this fashion instead of having participants
mark their agreement with a statement (as is typical in Likert scales) such
as ``The robot's disposition is positive.'' We did so because disagreeing with
that statement does not necessarily mean the same thing as the robot
having a ``very negative'' disposition: disagreeing with positive does not
imply agreeing with negative. This is important because here, we are just as interested
in whether timing can cause the perception of
negative disposition as we are in whether timing can cause the perception
of a positive one.

\prg{Subject Allocation} The experiment was within-subjects,
every participant saw each of the 4 conditions. There were a total
of 40 participants (80\% male and 20\% female, median age $=29$).
As in the exploratory study, all participants
were from the United States and were recruited through Amazon's
Mechanical Turk, with a minimum approval rating of 95\%.

\prg{Hypotheses} We state generic and intuitive hypotheses, motivated
in part by prior work findings when it comes to disposition and weight,
and extrapolating to confidence and competence. However, the devil is
in the details, and as we will see in the analysis, not all factors
will have their anticipated effects, nor the effect sizes will be the same.
Our mathematical models will leverage these details.

We hypothesize that faster motion is more positively perceived (it has
already been shown before to positively affect disposition-related
perceptions \cite{blythe1999motion}, and this could extrapolate), 
and makes objects look lighter (known from
animating dropping objects \cite{owen1999timing}):

\textbf{H1.} \emph{Increasing speed positively affects perceived
  competence, confidence, disposition, naturalness, and negatively
  affects perceived weight. }

In contrast, pausing (incorporating infinitesimally slow motion)
should have the opposite effect:

\textbf{H2.} \emph{Pausing negatively affects perceived competence,
  confidence, disposition, naturalness, and positively affects
  perceived weight.}

\subsubsection{Analysis}

We first performed a multivariate analysis on the data, and found that the
different items were not highly correlated (we computed item
reliability, and found Cronbach's $\alpha=.67$), so we proceeded with
separate analyses for each.

We used a factorial repeated measures ANOVA with speed and pause as
factors for each dependent measure. \figref{fig:results_exp2}
plots the results.

\prg{Competence} We found a significant main effect for \be{pause}
($F(1,163)=65.81$, $p<.0001$), and no other effects (main or
interaction). Pausing made the robot seem \emph{less}
competent. Surprisingly, moving faster made the robot seem only
ever-so-slightly more competent, suggesting that it is not overall
efficiency that matters for perceived competence.

\prg{Confidence} \be{Pausing} made the robot seem significantly
\emph{less} confident ($F(1,163)=45.60$, $p<.0001$). But unlike for
competence, higher \be{speed} made the robot seem significantly
\emph{more} confident ($F(1,163)=10.79$, $p=0013$), but resulted in a
smaller mean difference than pausing. The interaction effect was not
significant.

\prg{Disposition} \be{Pausing} resulted in a more \emph{negative}
disposition ($F(1,163)=24.08$, $p<.0001$). Surprisingly, speed did not
have a significant effect, though moving faster did result in a
sightly more positive perception, in line with prior work
\cite{blythe1999motion}.

\prg{Naturalness} Again, we found that \be{pausing} has a significant
main \emph{negative} effect ($F(1,163)=68.01$, $p<.0001$). Pausing
made the motion less natural, intuitively because it is not as smooth,
or because it does not match what a person would expect the robot to
do. Speed had a very marginal positive effect ($F(1,163)=1.85$,
$p=.1766$), though perhaps looking at other values for overall speed
would lead to the motion becoming less natural.

\begin{figure*}
  \includegraphics[width=\textwidth]{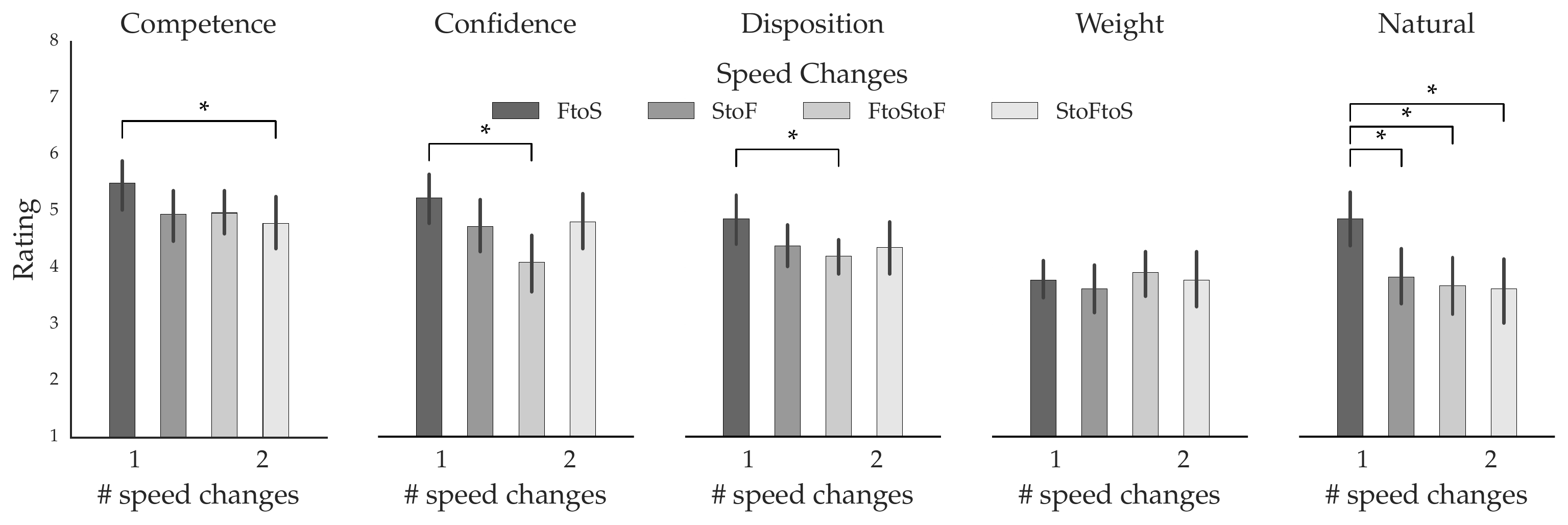}
  \caption{ Our second hypothesis-driven experiment measured the
    effect of speed changes on competence, confidence, disposition,
    weight, and naturalness.  Fast-to-Slow was the highest rated
    pattern for every property except weight.}
  \label{fig:results_exp1}
\end{figure*}
\prg{Weight} In the case of weight, it was \be{speed} that had a
significant \emph{negative} effect ($F(1,163)=19.97$, $p<.0001$), with
moving faster resulting in the object being perceived as lighter. This
is in line with animation advice for animating objects dropping, and
physically it makes sense that objects that drop faster are
lighter. But seeing this effect on a robot is important because
\emph{the object is no longer free, but rather being moved by an
  agent}. The robot does not need to move any different when the
object is heavier, and yet people do make inferences on weight based
on how the robot moves. Surprisingly, pausing did not affect weight,
even though pausing did make the robot seem less confident and
competent, which could suggest that it is carrying something heavier.

\prg{Summary} Overall, the effects we \emph{did} find were intuitive:
pausing negatively affected competence, confidence, disposition, and
naturalness, while speed positively affects confidence and negatively
affects weight. Participant comments suggested that
pauses make the robot look like it is ``planning'' -- it is uncertain about 
something or trying to locate something. We build on this uncertainty
idea in our model in the next section.

It is the effects that we did \emph{not} find
that were surprising. For instance, speed did not seem to influence
perceived competence, but influenced perceived confidence. 
Pausing did not seem to influence perceived weight. 
Of course, not finding an effect
does not mean it is not there, but here the means suggest a small
effect size, if at all. We dig deeper into these findings in our model
section.

\subsection{Speed Change Patterns}
\subsubsection{Experiment Design}
\prg{Manipulated Factors} We manipulated \be{speed changes} as in
\sref{sec:exploratory-design}, using the levels for 1 and 2 changes (previous
experiment already evaluated 0 changes).

\prg{Dependent Measures} We used the same measures as in \sref{sec:speedpause}.

\prg{Subject Allocation} There were 40 participants (59\% male and
41\% female, median age $=37$), selected and allocated as in \ref{sec:speedpause}.

\prg{Hypothesis} We hypothesize that changes in speed will make the
robot seem more hesitant and have a negative disposition, but make the
object look heavier:

\textbf{H3.} \emph{More changes in speed have a negative effect on
  perceived competence, confidence, disposition, naturalness, and a
  positive effect on perceived weight.}

Which kind of changes (e.g. \emph{StoF} vs \emph{FtoS})
have which effect remains to be determined.

\subsubsection{Analysis}

\prg{Number of Speed Changes} We first analyzed the effects that the
number of speed changes has, combining data from this experiment with
data from the former. A regression analysis shows, in line with our
hypothesis, that having \be{more changes} significantly \emph{decreases} perceived
competence ($F(1,163)=21.30$, $p<.0001$), confidence
($F(1,192)=12.24$,$p=0.006$), disposition ($F(1,163)=14.59$, $p=.0002$),
and naturalness ($F(1,163)=37.23$, $p<.0001$). It does \emph{not},
however, significantly affect perceived weight, and in fact the slope
on the linear fit is very close to $0$, namely $.02$. This is
consistent with our finding that pausing did not significantly affect
perceived weight, but counter-intuitive nonetheless.

\prg{Speed Change Patterns} Aside from number of changes, the actual
pattern is interesting as well -- does it make a difference, for
instance, if the robot starts slower and accelerates, or starts faster
and decelerates? We ran a repeated measures ANOVA for each dependent
measure, and found a significant effect for every case but perceived
weight, so we followed up with Tukey HSD. The results are plotted in
\figref{fig:results_exp1}.

For competence, we found that \emph{FtoS} was the best option,
significantly better than \emph{StoFtoS}, the worst option
($p=.0372$). This was similar for confidence, but here the worst
option was \emph{FtoStoF}. Disposition had the same result as
confidence. For naturalness, \emph{FtoS} was better than every other
option, all with $p<.03$.

\prg{Summary} Overall, more speed changes negatively impacted
all perceptions but weight.
Slowing down was the most positively perceived speed change of all.
At least for manipulation
tasks, if the robot is going to change speed, slowing down will make
it seem more competent, confident, and natural compared to speeding up or even
speeding up and then slowing back down. This is somewhat surprising, but
likely has to do with the notion of reaching a goal that the robot
needs to do something with, like handing over the bottle or picking it
up. Indeed, participants did often comment in this condition
that the robot is changing speed to hand the object over more 
smoothly.

Surprisingly, speed changes had no effect (close to 0 slope) on
perceived weight, even though intuitively the ability to change
speed could indicate a lighter object, and the need to
change speed could indicate a heavier object. Neither option
seemed to be the case.

%% file: models.tex
\section{Candidate Mathematical \\Models for Timing-Based\\ Human Inferences}
\label{sec-models}

\begin{figure*}
\centering
\includegraphics[width=\textwidth]{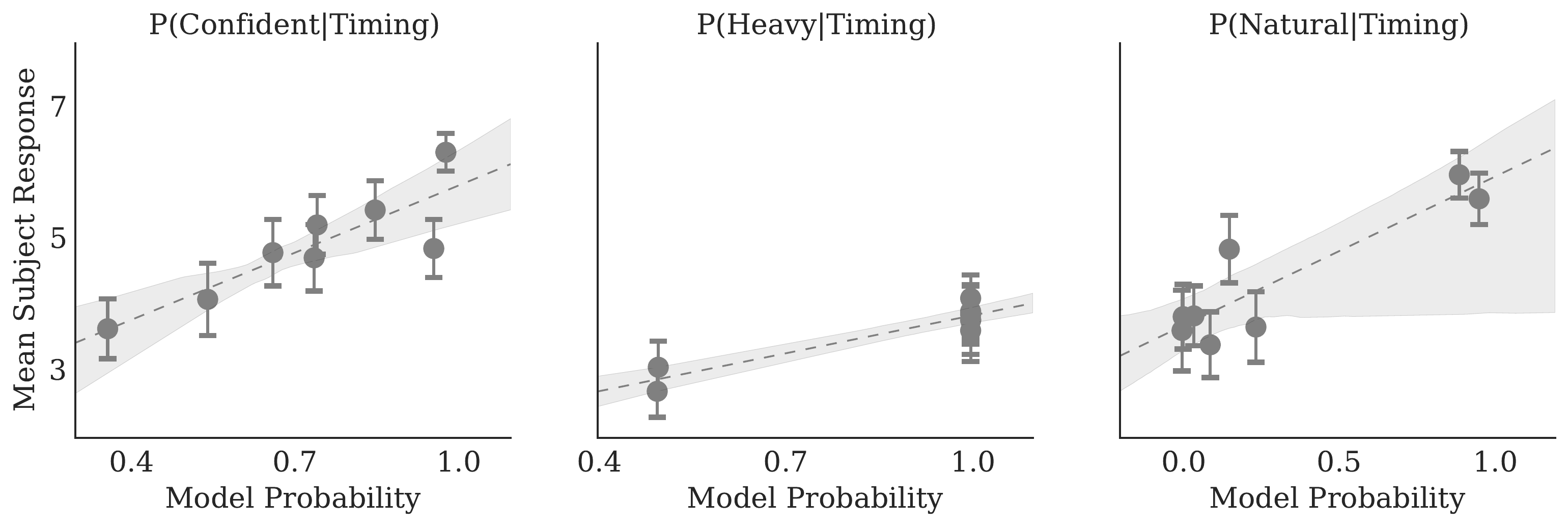}
 \caption[Correlation between our model predictions and real user
   data.  We varied the timing of the trajectories across 8 different
   conditions (described in Section~\ref{sec-experiment}) and the
   plots above have one data point per condition. The x-coordinate of
   each data point is the best model's prediction for that condition
   and the y-coordinate is the mean subject response, with a 95\%
   confidence interval.  For confidence and naturalness we show a 95\%
   confidence interval on the regression.  We see that higher
   probability for the robot being confident, the object being heavy,
   and the motion being natural usually does imply a higher rating
   along these criteria from the users. This suggests that the models
   are good candidates for capturing the inferences that people make,
   enabling robots to predict what their timing will convey. We also
   test how well the models can fit random data as opposed to real
   user data to check that they actually approximate the inferences
   that people make and not just overfitting.]{Correlation between our
   model predictions and real user data.  We varied the timing of the
   trajectories across 8 different conditions (described in
   Section~\ref{sec-experiment}) and the plots above have one data
   point per condition. The x-coordinate of each data point is the
   best model's prediction for that condition and the y-coordinate is
   the mean subject response, with a 95\% confidence interval. We also
   show a 95\% confidence interval on the regression. We see that
   higher probability for the robot being confident, the object being
   heavy, and the motion being natural usually does imply a higher
   rating along these criteria from the users. This suggests that the
   models are good candidates for capturing the inferences that people
   make, enabling robots to predict what their timing will convey. We
   also test how well the models can fit random data as opposed to
   real user data to check that they actually approximate the
   inferences that people make and not just overfitting.}
\label{fig:model-fit}
\end{figure*}

Our findings inform us about the effects of motion timing, but 
they are only \emph{descriptive} and not \emph{constructive}: 
the robot can't use them to time its motion
automatically in order to express what it wants. All the robot can do now is
compare the specific timings we explored and predict what users
will infer based on them. But what it needs instead is
to be able to predict how \emph{any} timing will be interpreted.

We take an inferential approach to enabling generalization in this 
section. We construct models of the inferences that people make
from robot motion timing based on our findings so far, and show
how they correlate with the real user data. Armed with such models,
the robot can \emph{simulate} what a new timing $\T$ would 
convey to a person, given a path, and even \emph{optimize} 
its timing to purposefully convey something.

\subsection{General Formulation}
We start with a general approach, and then fill in the details 
for confidence, weight, and naturalness.

We will model people's inference on some hidden robot or task state
$\theta$ (e.g., the robot's confidence) given timing $\T$ and path $\traj$
as evidence.
Thus, we model people as estimating $P(\theta|\T,\traj)$ via Bayesian inference 
from an observation model $P(\T|\traj,\theta)$. If the robot can approximate
the person's $P(\theta|\T,\traj)$, then it knows what $\T$ conveys about $\theta$. 


To model $P(\T|\traj,\theta)$, 
we suppose that the person expects the timing to be based on some criterion,
with different hidden variables leading to different criteria. We use $C(\T;\traj, \theta)$
to represent the criterion for a timing given a $\theta$ value
(e.g., a weight or a confidence value) and a path $\traj$.
Given $\theta$ and \traj, the probability of a trajectory timing
is
\begin{equation}
P( \T | \traj, \theta) \propto e^{-\lambda C(\T; \traj, \theta)} \label{eq-t-prob}
\end{equation}
Such a formulation has been used for paths and general actions in 
an MDP in \cite{dragan2013legibility, baker2009action, goodman2013knowledge, griffiths2008bayesian}.

In our experiments, the users observe timed trajectories and infer
$\theta$. To get this from our model, we apply Bayes' rule to compute
\begin{equation}
P(\theta | \T, \traj) \propto P(\T | \theta, \traj)P(\theta|\traj)
\end{equation}

Note that given this probability distribution, the robot can also search for
a timing for its path that maximizes the probability of a particular $\theta$, e.g. $\max_{\T} P(\theta_{1}|\T,\traj)$.

\prg{Model Evaluation and Parameter Selection} Next, we consider what
$C$ can be for confidence, weight, and naturalness. We will evaluate
these models by measuring the correlation between the model prediction,
for a given trajectory timing, with the mean subject response for that
timing. Our models include some free parameters (e.g., $\lambda$ in
\eref{eq-t-prob}) that we fit to the data by doing a grid search and
selecting the parameters that correlate best with the data. 

This means that there are two possible explanations for a high
correlation: either the model actually explains people's inferences,
or it is complex enough that it can overfit to any data. Thus, we have
a confound.  To address it, we run the same procedure on randomly
generated synthetic data: if we get a high correlation with random
data, then it is likely that our model has overfit. On the other hand,
a low correlation with random synthetic data suggests that our model
\emph{does actually help explain} the predictions that users made.

     
\subsection{Confidence}
\prg{Model} We observed that high speed led to an increase in
perceived confidence and pausing led to a decrease in perceived
confidence. We propose that a mathematical model for confidence can be the
precision (i.e., inverse variance) in the robot's belief state.

We thus model the observer as assuming, for simplicity, that the
robot's belief state is a Gaussian $\mathcal{N}(\mu,\sigma^{2})$ with
initial precision
$$\tau_0 = \frac{1}{\sigma_0^2}$$
where high $\tau_0$ corresponds to high
confidence and vice versa.

The robot gets observations at a constant
rate over the course of the trajectory. Our observer expects the
robot to use a different timing depending on the confidence --
intuitively, if it starts with low precision, it needs to get more
observations than if it starts with high precision.  More concretely,
the timing that the observer will expect for an initial precision,
$\tau_0$, is related to the cost $C$ that the observer expects the
robot to optimize when timing motion. We propose that $C$ should
target high final precision $\tau_{f}$, while trading off with being
efficient on the task:
\begin{equation}
C(\T; \traj, \tau_0) = k\duration + \frac{1}{\tau_f}
\end{equation}

If the robot moves faster, it gets fewer observations so
its final precision is lower. $k$ controls the relative importance of
speed versus precision. If each of these observations has Gaussian
noise with precision $\tau_{obs}$, then the robot's belief state
updates with a Kalman filter~\cite{kalman1960approach}. The precision at the end of the
trajectory is thus
\begin{equation}
\tau_{f} = n_{obs}\tau_{obs} + \tau_0
\end{equation}
where $n_{obs}$ is the number of observations the robot
gets during the trajectory. 

This first model attempt explains the interaction between
speed and perceived confidence, but can not explain the interaction
with pauses; paused trajectories in our experiment still have the same overall duration,
so the effect we found for pausing can not be explained by the current model.

To
account for pauses, we further suppose that the quality of each observation
depends on the robot's velocity. 
If the robot is not moving, then it gets observations with precision
$\tau_{obs}$. As the robot speeds up, the precision of its
observations decreases. This gives us the following formula for $\tau_f$:
\begin{equation}
\tau_f = \tau_0 + \sum_\ind \left(\dur_{\ind+1} - \ts\right)  \frac{\tau_{obs}}{1 + r\norm{\vel}}.
\end{equation}
Recall that \ts{} is the time the robot reaches configuration \traji{} and
\vel{} is the corresponding velocity. $r$ governs how quickly the
observation precision falls off as the robot speeds up.

The inference task is to determine the value of $\tau_0$, given a
timed trajectory. We consider two possible values for $\tau_0$:
$\tau_0=1$ represents ``high confidence'' and $\tau_0=0.5$ represents
``low confidence.''

\prg{Evaluation} We used grid search to fit $r, k$ and $\lambda$. For
each parameter we consider 10 values between $10^{-2}$ and $10^2$,
evenly distributed in log space. The best fit parameters were $r=10^2,
k=0.6, \lambda=12.9$. The corresponding correlation is $0.86$. The
average best-fit correlation with random data was
0.3. \figref{fig:model-fit} (left) plots the confidence model's output
versus the mean student ratings for confidence.

\subsection{Weight}
\prg{Model} Previous work has shown that humans make inferences about
the weights of objects based on their motions and that their
perceptions can be modelled as Bayesian inference with a simplified
physics model~\cite{sanborn2009bayesian,
  hamrick2016inferring}. However, this work focused on objects in free
fall and collisions, while we are interested in objects being moved by
a different entity, namely the robot. We found that human inferences
about weight depend primarily on the speed of the object. A higher
speed led users to infer that the held object was lighter.

It is tempting to apply a
model where trajectories trade off between, e.g., energy for the robot (sum of squared
torques on its joints) and the duration of the trajectory. This would lead to the
appropriate inference with respect to speed; a higher weight means
that the same torque results in a lower speed. However a sum of
squared torques cost does not give a good explanation of the impact
speed changes or pauses had on the inferred weight. A robot minimizing
sum of squared torques will pay a higher penalty to pause with a heavy
object, so pausing would change the inferred weight in this model.

As an alternative, we model the robot as attempting to control the
overall \emph{momentum} of the object it is holding. In this model,
the robot is not minimizing the effort it expends to move the object,
but rather it minimizes the amount of effort it would take to bring
the object to a halt. The overall cost function trades off between
duration and the sum of momentums across the trajectory:
\begin{equation}
C(\T;\traj, m) = k\duration + m\sum_i \norm{\velEE}
\end{equation}
where $m$ is the mass of the object being held, $\duration$ is the duration,
 and $\velEE$ is the velocity of the end effector.
  In this model, the inferred mass of the object will depend
on the average velocity of the object and does not have any dependence
on speed changes that occur during the trajectory. This is in
contrast to cost functions that minimize the sum of squared torques or
the kinetic energy of the object. 

The inference task is to determine $m$, given a timed trajectory. We
considered two physically plausible values of $m$: $m=0.5$kg
represents a light mass and $m=0.8$kg represents a heavy mass.

\prg{Evaluation} We used grid search to fit $k$ and $\lambda$. For
each parameter we considered 10 values between $10^{-2}$ and $10^2$,
evenly distributed in log space. The best fit parameters were $k=4.6,
\lambda=35.9$. The corresponding correlation is $0.93$. The average
best-fit correlation with random data was 0.18. \figref{fig:model-fit}
(center) plots the confidence model's output versus the mean student
ratings for confidence.

\subsection{Naturalness}
\prg{Model} Our model for naturalness is the simplest of the
three. Natural human arm motions can be modeled as minimizing an
objective function defined as the magnitude of jerk integrated over the
motion \cite{flash1985coordination}.
In the case of robot-human handovers (the task used in our experiment),
it has been shown that minimum jerk motions lead to faster reaction
times from the human \cite{huber2008human}.

The cost function for naturalness inference is therefore a tradeoff between the
duration of the trajectory (as before) and the sum of squared jerks
along the trajectory:
\begin{equation}
C(\T;\traj, k) = k\duration + \sum_i ||J_i||^2
\end{equation}
where $k$ is the naturalness parameter that governs how natural the
trajectory should be in comparison to its duration. $J_i$ is the jerk
associated with the $i^{th}$ we can express this in terms of stepwise
velocities as
\begin{equation}
J_i = v_{i+1} + v_{i-1} - 2v_i
\end{equation}

The inference task is to determine $k$ given a trajectory timing. We
suppose that $k$ can take on two possible value $k_{high}$ and
$k_{low}$ that we fit to the data.

\prg{Evaluation} We used grid search to fit $k_{high}, k_{low},
\text{and }\lambda$. We considered 10 values between $10^{-2}$ and
$10^2$, evenly distributed in log space. During the grid search, we
enforced the constraint that $k_{high} > k_{low}$. The best fit was
$k_{high} = 100, k_{low}=1.66, \lambda=4.64$. The corresponding
correlation was $0.90$. The average best-fit correlation with random
data was $0.29$. \figref{fig:model-fit} (right) plots the naturalness
model's output versus the mean student ratings for naturalness.


%% file: discussion.tex
\section{Discussion}

\prg{Summary} We already knew from prior work that timing is
important, and expected to see effects on perceptions of
non-functional properties of the robot, like disposition and
naturalness. More exciting is that we have also found effects on
perceptions of functional properties as well, like competence,
capability, and carried object weight.

We introduced mathematical models for some of these perceptions, whose
predictions strongly correlated with the perceptions of actual
users. These contribute to enabling robots to anticipate what their timing will
convey, as well as to optimize their timing, given a path, to
purposefully convey that they are not confident, that they are handing
the person over a heavy object, or to simply produce more natural or
predictable motion.

\prg{Limitations and Future Work} Despite these promising results
supporting the importance of timing and bringing us closer to
autonomous expressive timing, we have just scratched the surface of
this deep area of research.  Timing is complex and multi-faceted, and
we have only studied three factors that contribute to timing: speed,
changes of speed (in particular ways), and pausing (at particular
times).

Our models for weight, confidence, and naturalness help
generalize to new timings outside of the conditions in our study.
But more investigation is needed to put each model to the
test with novel timing situations, new paths, new
robots, and new tasks. Further, the fact that the current models correlate with the data we collected
does not necessarily imply that they produce useful timings when optimized. 
Performing the timing generation and adjusting the models accordingly 
is our main direction of future work. 

Finally, for each of our current models we defined a timing cost function based on
some physical or informational quantity (e.g., momentum in the weight
cost or precision in the confidence cost). Doing the analogous for effects like disposition
is a significant future challenge, because such quantities are hard to
directly relate to concrete physical properties. 

\section{Acknowledgments}
This work is supported by The Center for Information Technology Research
in the Interest of Society, Berkeley DeepDrive, and the Center for
Human-Compatible AI.